\def\year{2022}\relax
\documentclass[letterpaper]{article} 
\usepackage{aaai22}  
\usepackage{times}  
\usepackage{helvet}  
\usepackage{courier}  
\usepackage[hyphens]{url}  
\usepackage{graphicx} 
\usepackage{natbib}  
\usepackage{caption} 
\DeclareCaptionStyle{ruled}{labelfont=normalfont,labelsep=colon,strut=off} 
\usepackage{algorithm}
\usepackage{algorithmic}
\usepackage{pifont}
\usepackage{soul}
\usepackage{pifont}
\usepackage{bbding}
\usepackage{adjustbox}
\usepackage{colortbl}
\usepackage{amsmath}
\usepackage{amsfonts}
\usepackage{wasysym}
\usepackage{multirow}
\usepackage{booktabs}
\usepackage{subcaption}
\def\etal{\emph{et al.}}

\definecolor{Gray}{gray}{0.9}

\usepackage{xcolor}
\definecolor{bb}{rgb}{0.0, 0.0, 0.5}
\usepackage[colorlinks=true,linkcolor=red,citecolor=bb,urlcolor=red]{hyperref}

\newcommand{\ie}{\textit{i.e., }}
\newcommand{\eg}{\textit{e.g., }}

\usepackage{lipsum}
\usepackage{fixltx2e}
\usepackage{romannum}
\usepackage{newfloat}
\usepackage{listings}
\floatstyle{ruled}
\newfloat{listing}{tb}{lst}{}
\floatname{listing}{Listing}
\title{FINO: Flow-based Joint Image and Noise Model}

\author{
    Lanqing~Guo$^{1}$,$\;$~Siyu~Huang$^{1}$,$\;$~Haosen~Liu$^{2}$~$\;$and~$\;$Bihan Wen$^{1}$}
 \affiliations{
  $^{1}$School of Electrical \& Electronic Engineering, Nanyang Technological University, Singapore \\ $^{2}$SmartMore Corporation, China \\
  \texttt{\{lanqing001, siyu.huang, bihan.wen\}@ntu.edu.sg} $\quad\quad\quad$  \texttt{haosen.liu0803@gmail.com} \\
}
\usepackage{bibentry}

\begin{document}

\maketitle

\begin{abstract}
One of the fundamental challenges in image restoration is denoising, where the objective is to estimate the clean image from its noisy measurements. To tackle such an ill-posed inverse problem, the existing denoising approaches generally focus on exploiting effective natural image priors. The utilization and analysis of the noise model are often ignored, although the noise model can provide complementary information to the denoising algorithms. 
In this paper, we propose a novel Flow-based joint Image and NOise model (FINO) that distinctly decouples the image and noise in the latent space and losslessly reconstructs them via a series of invertible transformations. We further present a variable swapping strategy to align structural information in images and a noise correlation matrix to constrain the noise based on spatially minimized correlation information.
Experimental results demonstrate FINO's capacity to remove both synthetic additive white Gaussian noise (AWGN) and real noise.
Furthermore, the generalization of FINO to the removal of spatially variant noise and noise with inaccurate estimation surpasses that of the popular and state-of-the-art methods by large margins.

\end{abstract}

\section{Introduction}
Image denoising refers to recovering the underlying clean image from an observed noisy measurement. Despite today's vast improvement in camera sensors, digital images are often corrupted by severe noises in complex environments, resulting in nontrivial effects to subsequent vision tasks.

Existing image denoising methods generally rely on the construction of effective image priors. For conventional methods, the corresponding priors include, \eg sparsity~\cite{elad2006image, ravishankar2012learning} and low-rankness~\cite{gu2014weighted}.
By shrinkage or filtering in the transform domain, image components that are satisfied with the prior are preserved in the denoised results. 
However, these priors are usually applied to image patches, lacking patch consensus and global modeling.
Recently, the deep learning approaches~\cite{zhang2017beyond,zhang2018ffdnet,liu2018non} 
have achieved state-of-the-art image denoising results by relying on an external training corpus. 
These deep denoisers generally learn a direct mapping from noisy images to clean images, where the learned models serve as an effective prior on the clean image space.
Despite the success of learning effective image priors for denoising, few work to-date investigated the noise modeling that is complementary to the deep image priors learning.
In practice, the residual between the noisy image and the denoised one always contains image structures that are wrongly removed together with noise. These structures generally correspond to high-frequency components of an image, which are distinct from the random noise. This inspires us that it might be possible to constrain the residual map with the noise model to `squeezing' image information from the residual map.

\begin{figure}[t]
\centering
\includegraphics[width=\linewidth]{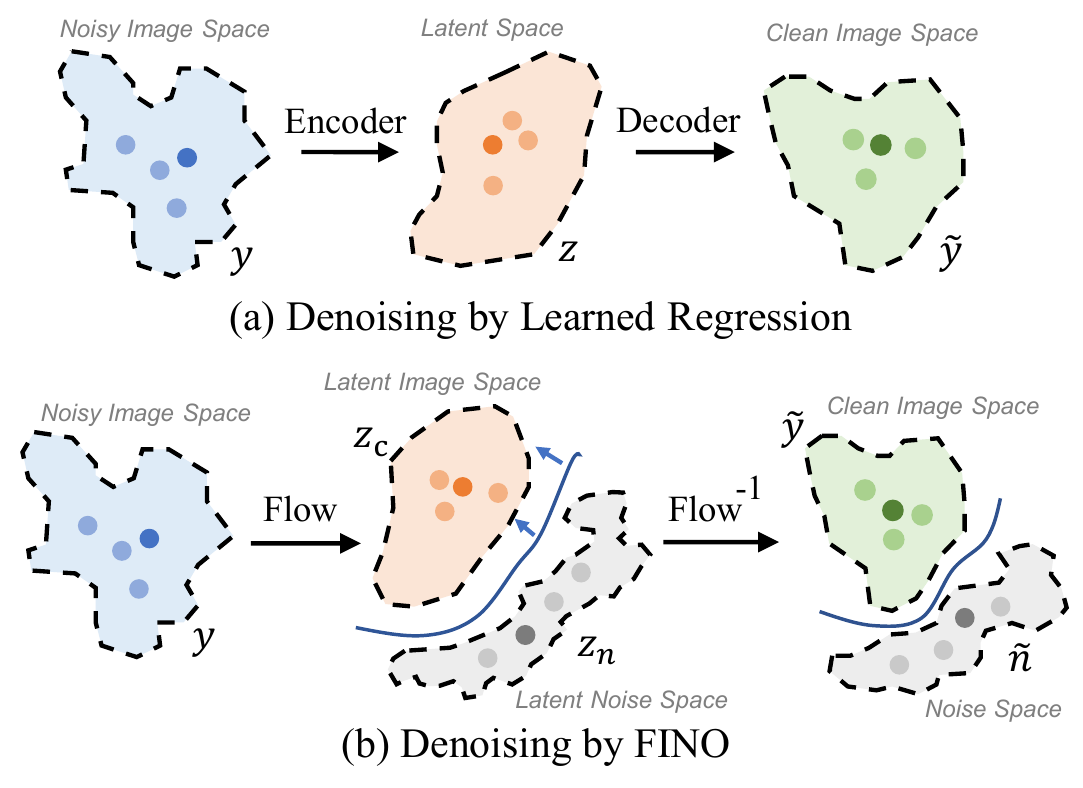}
\vspace{-0.6cm}
\caption{
The framework of Flow-based Joint Image and Noise Model (FINO) and comparison between previous regression-based denoisers (a) and the proposed FINO (b).
}
\label{fig:motivation}
\end{figure}

Different from all existing deep denoising methods, we reformulate the image denoising task as a dual modeling problem of both the image and noise (as illustrated in Figure~\ref{fig:motivation}(b)), instead of only reconstructing the noise-free image (Figure~\ref{fig:motivation}(a)). 
However, \textit{how to achieve an effective decoupling of clean image and noise} is still a challenging and under-explored problem in the existing literature.
In this paper, we show that the noisy images can be losslessly transformed to a more distinguishable feature space through a novelly proposed framework named \textbf{F}low-based joint \textbf{I}mage and \textbf{NO}ise model (\textbf{FINO}). FINO  losslessly decouples the image-noise components in the latent space through a forward process of the flow-based invertible network. Then, the decoupled components can be reconstructed as the noise and image in the spatial domain through a backward process of an invertible network.
Based on the decoupled noise and image components,
we further introduce a noise variable swapping strategy to align the structural information in images, as well as a constraint on the noise correlation matrix to be spatially independent on the neighboring regions.
Extensive experiments are conducted to evaluate FINO on both the synthetic noise and real noise removal tasks. Empirical results show that FINO achieves superior performances in comparison with the state-of-the-art image denoising methods. Besides, FINO provides significantly better generalizability than the existing denoising methods.

The contributions of this work are summarized as follow:
\begin{itemize}
    \item We propose to jointly model the distributions of the image and noise for denoising tasks, showing that the noise model can provide abundant and complementary information, in addition to image priors. 
    
    \item We present a novel image denoising framework named FINO which distinctly disentangles the noise and the noise-free image in the latent space. Two learning methods, including variable swapping and noise correlation matrix, are also proposed to improve the learning of FINO.
    
    
    \item We conduct extensive experiments on both the synthetic and real noise datasets. FINO shows superior denoising and generalization performances compared to the existing denoising methods.
\end{itemize}

\section{Related Work}

\subsection{Image Denoising}
\vspace{1mm}
\noindent
\textbf{Model-based image denoising.}
Image denoising is a typical ill-posed problem with the goal of recovering high-quality images from their noisy measurements. Numerous efforts have been made towards it over the past decades. Classic methods generally take advantage of the image priors, such as sparsity~\cite{elad2006image, ravishankar2012learning}, low rank~\cite{gu2014weighted}, and non-local self-similarity~\cite{dabov2007color,xu2015patch}, to address the denoising problem.
Most classic denoisers utilize the image features in certain transform domains by applying shrinkages or filtering to the exploitation of image priors.
The learning-based transform methods are more flexible, but they all worked on patches, lacking global modeling.
Motivated by this, in this work, we employ a flow-based invertible network to conduct a learnable global transforming. 

\vspace{1mm}
\noindent
\textbf{Deep learning based image denoising.}
In recent years, deep learning-based denoisers exhibit superiority in learning the end-to-end mapping from noisy to clean images~\cite{liu2018non}. 
For instance, Zhang~\etal~\cite{zhang2017beyond} achieves a very competitive denoising performance through residual neural networks.
Zhang~\etal~\cite{zhang2018ffdnet} further introduces a noise level map to control the trade-off between noise reduction and detail preservation.
More recently, some researchers~\cite{Ploetz:2018:NNN,liu2018non} try to leverage deeper and larger neural networks to achieve better performance.
Generally, existing denoising methods focus on exploiting effective natural image priors, while the modeling, analysis, and utilization of the noise component are often ignored.
This work jointly models the natural image and noise via invertible neural networks to deliver better denoising performance and visual quality.

More recently, there are some attempts~\cite{yue2019variational,anwar2019real,guo2019toward} for denoising on real-noisy images. The attempts can be generally divided into two categories: 1) Two-step denoising~\cite{yue2019variational}, which first estimates the noise map then reconstructs the clean image non-blindly based on the estimated noise map.
2) One-step denoising with an end-to-end framework~\cite{anwar2019real}.
This work focuses on both real noise and synthetic noise removal. It follows the one-step denoising approaches to enhance the generalization capacity of denoising models.

\subsection{Neural Flows}
The neural flow is a type of deep generative model that learns the exact likelihood of targets through a chain of reversible transformations.
The generative process $\mathbf{x} =\mathcal{F}_{\boldsymbol{\theta}}(\mathbf{z})$ given a latent variable $\mathbf{z}$ can be specified by an invertible architecture $\mathcal{F}_{\boldsymbol{\theta}}$.
The direct access to the inverse mapping is $\mathbf{z} = \mathcal{F}_{\boldsymbol{\theta}}^{-1}(\mathbf{x})$.
As a pioneering work, NICE~\cite{dinh2014nice} learns a highly non-linear bijective transformation that maps the training data to a space where its distribution is factorized.
Following NICE, more effective and flexible transformations have been proposed~\cite{dinh2016density,ho2019flow++,NEURIPS2018_d139db6a}.

More recently, a series of works exploit neural flows for image restoration~\cite{abdelhamed2019noise,lugmayr2020srflow,xiao2020invertible,liu2021invertible}, which formulate image restoration as a non-degradation image generation problem. For instance, SRFlow~\cite{lugmayr2020srflow} designs a conditional normalizing flow architecture for super-resolution, which learns the distribution of realistic HR images.
\cite{xiao2020invertible} and \cite{liu2021invertible} apply invertible networks to image rescaling and image denoising tasks, respectively.
InvDN~\cite{liu2021invertible} focuses on real noise removal and detours the noise-image disentanglement.
It splits noisy images into low-frequency and high-frequency components and then directly drops the high-frequency component, restoring the image based on the low-frequency component only, which may result in information loss and over-smoothness.
Different from InvDN, our proposed FINO utilizes an invertible network to decouple the image content and noise components in the latent space and then reconstruct them in the image space, respectively, achieving a lossless image-noise disentanglement.


\section{Preliminary}
\begin{figure}[t]
\centering
\includegraphics[width=\linewidth]{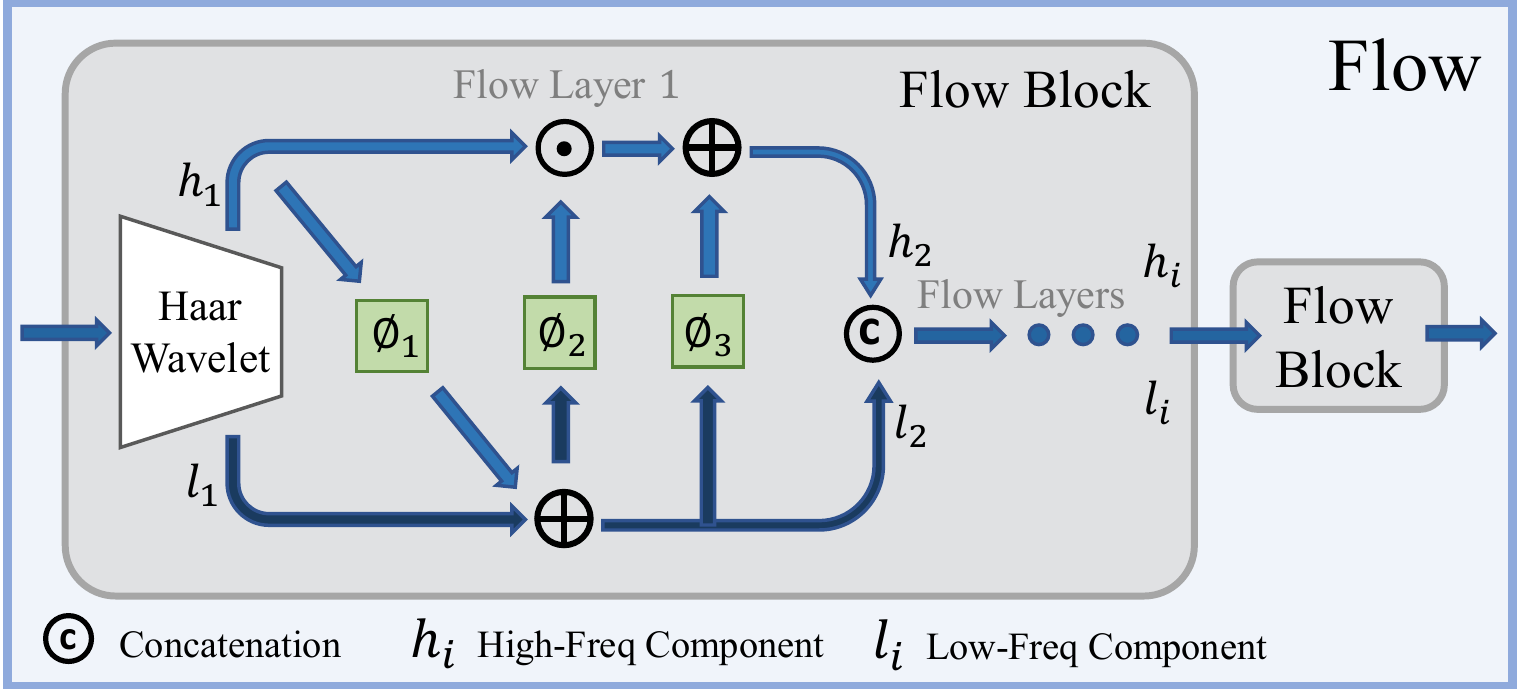}
\vspace{-0.3cm}
\caption{
The architecture of neural flow model, consisting of two flow blocks. Each flow block includes a invertible Haar wavelet transformation at first layer and twelve affine coupling layers. $i$ denotes $i$-th affine coupling layer.
}
\label{fig:inn}
\end{figure}

In this section, we introduce the preliminary knowledge of the multi-scale neural flow~\cite{ardizzone2019guided,xiao2020invertible,liu2021invertible}. We denote it as $\mathrm{Flow}(\cdot)$ in this paper. As shown in Figure~\ref{fig:inn}, $\mathrm{Flow}(\cdot)$ consists of a series of flow blocks, and each flow block consists of an invertible wavelet transformation followed by a series of affine coupling layers.

\vspace{1mm}
\noindent\textbf{Invertible wavelet transform.} 
To disentangle the information of clean image and noise, we employ invertible Haar wavelet transformation at the first layer of each flow block to downsample the input images/features and to increase the feature channels~\cite{xiao2020invertible}.
After the wavelet transformation, the input image/features with a shape of $(H,W,C)$ should be squeezed into $(H/2, W/2, 4C)$. $4C$ denotes three directions of high-frequency coefficients and one low-frequency representation~\cite{kingsbury1998wavelet}. The invertible wavelet transformation provides the separated low and high-frequency information to the following invertible neural layers. 

\vspace{1mm}
\noindent\textbf{Affine coupling layers.}  
After the wavelet transformation layer, the input image/feature $\mathbf{u}_i$ has been splitted into low and high-frequency components, denoted as $\mathbf{h}_i$ and $\mathbf{l}_i$, respectively. We leverage the coupling layer~\cite{dinh2014nice} to further decouple the structural information and the degradation bias. Suppose the $i$-th coupling layer's input is $\mathbf{u}_i$ and the output is $\mathbf{u}_{i+1}$ $(i=1\ldots I)$, the forward procedure in this block is
\begin{align}
    \nonumber \mathbf{l}_i,\mathbf{h}_i &= \mathrm{Split}(\mathbf{u}_i) \; , \\
   \nonumber \mathbf{l}_{i+1} &= \mathbf{l}_i + \phi_1(\mathbf{h}_i) \;, \\
   \nonumber \mathbf{h}_{i+1} &= \phi_2(\mathbf{l}_{i+1})\odot \mathbf{h}_i + \phi_3(\mathbf{l}_{i+1})\;, \\
    \mathbf{u}_{i+1} &= \mathrm{Concat}(\mathbf{l}_{i+1}, \mathbf{h}_{i+1})\;,
\end{align}
where $\mathrm{Split}(\cdot)$ denotes channel-wise splitting and $\mathrm{Concat}(\cdot)$ is the corresponding inverse operation. $\phi_1(\cdot)$, $\phi_2(\cdot)$, and $\phi_3(\cdot)$ can be any neural networks that are not required to be invertible. The backward procedure is easily derived as
\begin{align}
    \nonumber \mathbf{l}_{i+1},\mathbf{h}_{i+1} &= \mathrm{Split}(\mathbf{u}_{i+1}) \; , \\
       \nonumber \mathbf{h}_{i} &= (\mathbf{h}_i - \phi_3(\mathbf{l}_{i+1}))/\phi_2(\mathbf{l}_{i+1})\;, \\
   \nonumber \mathbf{l}_{i} &= (\mathbf{l}_{i+1} - \phi_1(\mathbf{h}_i))/\phi_1(\mathbf{h}_i) \;, \\
    \mathbf{u}_{i} &= \mathrm{Concat}(\mathbf{l}_{i}, \mathbf{h}_{i})\;.
\end{align}

\section{FINO}

In this section, we present a neural flow-based framework named FINO to jointly model the image and noise in context of image denoising. We first formulate the image denoising problem, then discuss why employing the invertible neural networks for image-noise disentangling, and introduce the architecture details of FINO, including the variable swapping in latent space for disentanglement, the clean image regression for image modeling, and the noise correlation matrix for noise modeling, as well as objective functions.

\begin{figure*}[t]
\centering
\includegraphics[width=.95\linewidth]{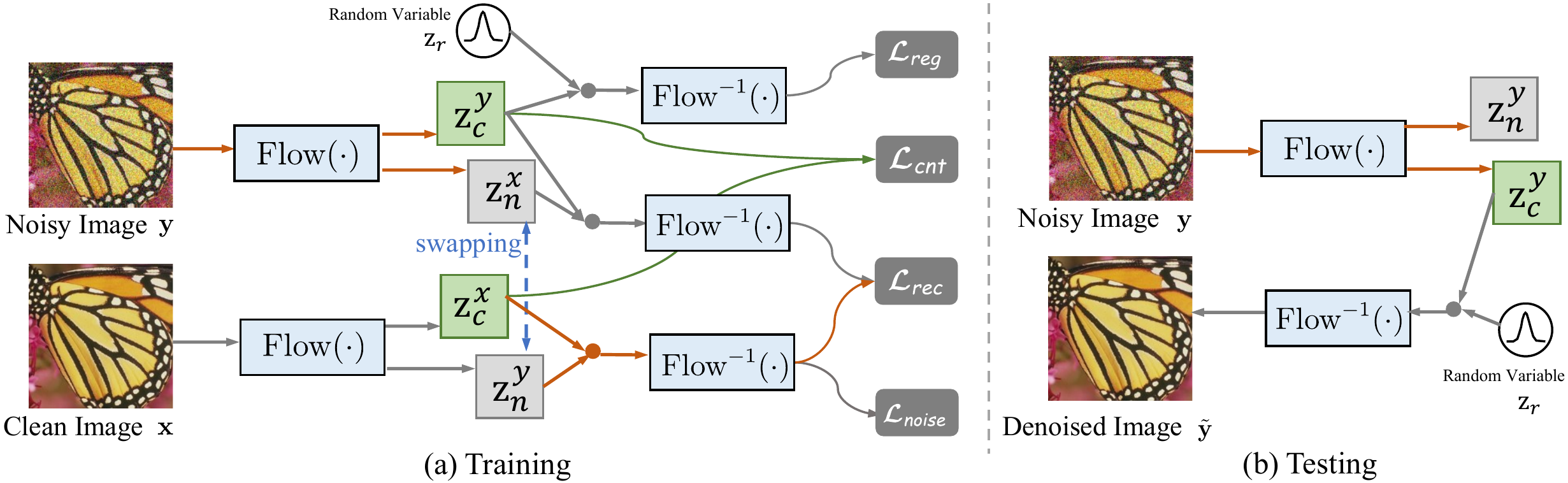}
\caption{
An overview of our Flow-based Joint Image and NOise Model (FINO).
}
\label{fig:pipeline}
\end{figure*}

\subsection{Problem Formulation}
A noisy image $\mathbf{y}$ and its noise-free counterpart $\mathbf{x}$ can be fromulated as
\begin{equation}
    \textbf{y}=\textbf{x}+\textbf{n}\;,
\end{equation}
where $\mathbf{n}$ denotes the random noise.
Here, we consider both the synthetic noise and the real noise. The synthetic noise is the \textit{i.i.d} additive white Gaussian noise (AWGN). It follows the normal distribution $\mathcal{N}(0,  \sigma^2 \cdot \mathbf{I})$.
Typical supervised regression-based denoising methods train the deep neural networks (DNNs) by 
\begin{equation}
\min _{\boldsymbol{\theta}} \; \mathbb{E}_{\textbf{x}, \textbf{y}} [\mathcal{L}_{reg}\left(\mathcal{G}_{\boldsymbol{\theta}}(\textbf{y}), \textbf{x}\right)]\;,
\end{equation}
where $\mathcal{G}_{\boldsymbol{\theta}}(\cdot)$ denotes the regression-based denoiser, which can be regarded as the combination of encoder and decoder $\mathcal{D}_{\boldsymbol{\theta}}(\mathcal{E}_{\boldsymbol{\theta}}(\cdot))$. $\mathcal{L}_{reg}(\cdot,\cdot)$ denotes the loss function, \eg the $\ell_1$ or $\ell_2$ loss.

Different from the conventional methods, we novelly propose a dual modeling of both noise and image as 
\begin{align}
\min _{\boldsymbol{\theta}} \;\mathbb{E}_{\textbf{x}, \textbf{y}} [ \mathcal{L}_{reg}\left(\mathcal{G}_{\boldsymbol{\theta}}(\mathbf{y}), \textbf{x}\right) +
\mathcal{L}_{noise}\left(\mathcal{H}_{\boldsymbol{\theta}}(\mathbf{y}), \textbf{n}\right)]\;,
\end{align}
where $\mathcal{L}_{noise}(\cdot, \cdot)$ denotes the loss function of the noise modeling, and $\mathcal{H}_{\boldsymbol{\theta}}(\cdot)$ denotes the noise model.
The proposed FINO disentangles noise and image in a more distinguishable latent space via an invertible network, on behalf of $\mathcal{G}_{\boldsymbol{\theta}}$ and $\mathcal{H}_{\boldsymbol{\theta}}$, by jointly modeling image and noise.


\subsection{Why Disentangling via Invertible Network?}
Although it is known that the representation ability of the invertible network is limited and weaker than some more sophisticated deep neural networks, because of its specially designed structure~\cite{kirichenko2020normalizing}, there are three main reasons why we choose the invertible network for image-noise decoupling.

\begin{itemize}
\item Firstly, existing regression-based image denoising algorithms cannot achieve lossless image reconstruction of the input image: Once the input noisy image is embedded into latent space, some information may get lost. This problem is especially severe for the non-structural information, \eg noise.
Different from them, the invertible network can losslessly encode the image and noise. This property ensures that recoveries of the image and noise can always complement each other, providing the basis of improving the image denoising performance via a joint image and noise modeling.

\item Secondly, only the forward module of invertible network needs to be trained, while the backward module is the direct inverse, which respectively act as encoder and decoder. The number of free parameters can be thus significantly reduced. 


\item Finally, image denoising is an ill-posed inverse problem of one-to-many mapping, \ie one noisy image can be restored to many denoised estimations. Most of the existing methods formulate it as a one-to-one mapping task, \ie delivering one denoised image from one noisy input. However, FINO can sample diverse denoised images by coupling the disentangled clean image with any random variable sampled from a normal distribution.
\end{itemize}

\subsection{The Framework of FINO}\label{sec:fino}

Given a noisy image $\textbf{y}$ and its corresponding clean counterpart $\textbf{x}$ in training stage, we first employ an invertible flow model $\mathrm{Flow}(\cdot)$ to embed the input pair to latent codes respectively, as
\begin{equation}
    \textbf{z}^{x} = \mathrm{Flow}(\textbf{x}) \qquad \textbf{z}^{y} = \mathrm{Flow}(\textbf{y})\;.
\end{equation}

\vspace{1mm}
\noindent\textbf{Variable swapping for disentangling.}
We divide the latent space $\mathcal{Z}$ into two sub-spaces, \ie clean image space and noise space, with separated latent codes $\textbf{z}=[\textbf{z}_c, \textbf{z}_n]$. 
As shown in Figure~\ref{fig:pipeline}(a), in order to ensure the noise information is decoupled from the noisy inputs, we introduce a noise variable swapping strategy to generate noisy image,
 \ie combining the noise variable $\textbf{z}_n^y$ from noisy input and the clean image variable $\textbf{z}_c^x$ from the clean counterpart as follows,
\begin{equation}
\label{eq:flow1}
    \hat{\mathbf{y}} = \mathrm{Flow}^{-1}(\textbf{z}_c^x, \textbf{z}_n^y) \qquad \hat{\mathbf{x}} = \mathrm{Flow}^{-1}(\textbf{z}_c^y, \textbf{z}_n^x)\;,
\end{equation}
where $\hat{\textbf{y}}$ is the reconstructed noisy image, $\hat{\textbf{x}}$ is the reconstructed clean image, and $\mathrm{Flow}^{-1}(\cdot,\cdot)$ denotes the inverse process of $\mathrm{Flow}(\cdot)$. Following Equation \ref{eq:flow1}, we can easily derive the reconstructed noise $\hat{\textbf{n}} = \hat{\textbf{y}} - \textbf{x}$.

To ensure noise information is completely embedded in the noise variable $\textbf{z}_n$, we impose a \textit{reconstruction loss} to 
encourage recovery of the original noise $\mathbf{n}$ and clean image $\mathbf{x}$, using the noise and clean image variables, respectively:
\begin{equation}
\mathcal{L}_{rec}=\left\|\hat{\mathbf{n}} - \textbf{n}\right\|_1 + \left\|\hat{\mathbf{x}} - \textbf{x}\right\|_1 \;.
\end{equation}

Besides, to enforce the clean image code $\textbf{z}_c$ to contain only the noise-free content information, we employ a \textit{content alignment loss} to align the structural features of noisy and clean image pairs, as
\begin{equation}
\mathcal{L}_{cnt}=\left\|\mathbf{z}_c^x-\mathbf{z}_{c}^y\right\|_1\;.
\end{equation}

\vspace{1mm}
\noindent\textbf{Clean image regression.}
In addition, the denoised images can also be generated via its disentangled clean image component $\textbf{z}_c^y$ and a random variable $\textbf{z}_r$ as follow
\begin{equation}
    \Tilde{\mathbf{y}} = \mathrm{Flow}^{-1}(\textbf{z}_c^y, \textbf{z}_r)\;.
\end{equation}
We employ a \textit{regression loss} on the generated $\Tilde{\mathbf{y}}$ and further enforce the noise variable to be independent of the structure information, as
\begin{equation}
\mathcal{L}_{reg}=\left\|\Tilde{\mathbf{y}} - \textbf{x}\right\|_1\;.
\end{equation}

Besides, the restored image can be sampled by combing their internal clean image variable and a normal distribution variable in the testing stage as shown in Figure~\ref{fig:pipeline}(b).

\begin{table*}[!t]
\centering
\footnotesize
\setlength{\tabcolsep}{0.4em}
\caption{Quantitative results of denoised PSNR (in dB)$\uparrow$ on CBSD68~\cite{roth2005fields}, Kodak24~\cite{franzen1999kodak}, and Set5~\cite{BMVC.26.135}. The baseline methods include CBM3D~\cite{dabov2007color}, CDnCNN~\cite{zhang2017beyond}, and FFDNet~\cite{zhang2018ffdnet}. In each column, the best result is highlighted in
\textbf{bold}.}
\adjustbox{width=1.\linewidth}{
    \begin{tabular}{c|l|ccc|ccc|ccc|ccc|ccc}
        \toprule
               \multirow{2}{*}{Datasets} &  estimated $\sigma$ & \multicolumn{3}{c|}{$\sigma =15$} & \multicolumn{3}{c|}{$\sigma =25$} & \multicolumn{3}{c|}{$\sigma =35$} & \multicolumn{3}{c|}{$\sigma =50$} & \multicolumn{3}{c}{$\sigma =75$}\\
               & testing $\sigma$ & $10$ & $15$ & $20$& $20$ & $25$ & $30$& $30$ & $35$ & $40$ & $45$ & $50$ & $55$ & $70$ & $75$ & $80$\\\midrule
        \multirow{4}{*}{CBSD68} & CBM3D & 34.68 &33.52& 29.51 & 31.27 & 30.71 & 28.69 & 29.33 & 28.89 & 27.54 & 27.53& 27.38& 26.93 & 25.87 & 25.74& 25.45\\
        & CDnCNN & 34.68 &33.89 &29.06  &31.68 &31.23 & 28.50 & 29.81 &29.58 &27.63 & 28.05 &27.92 &26.83 & 24.35 &24.47 &24.44 \\
        & FFDNet & 34.60 &33.87 &30.05 & 31.48 &31.21 &29.09 & 29.70 &29.58 &28.27 & 28.00 &27.96 &27.22 & 26.26 &26.24 &25.89 \\
         & \cellcolor{Gray}FINO & \cellcolor{Gray}\textbf{34.96} &\cellcolor{Gray}\textbf{34.05} &\cellcolor{Gray} \textbf{30.81} &\cellcolor{Gray}\textbf{31.82} &\cellcolor{Gray}\textbf{31.43} &\cellcolor{Gray}\textbf{29.52} & \cellcolor{Gray} \textbf{29.96} &  \cellcolor{Gray}\textbf{29.76} & \cellcolor{Gray} \textbf{28.45} & \cellcolor{Gray}\textbf{28.24} &\cellcolor{Gray}\textbf{28.21} &\cellcolor{Gray}\textbf{27.34} & \cellcolor{Gray}\textbf{26.49} & \cellcolor{Gray}\textbf{26.45} &\cellcolor{Gray}\textbf{26.08} \\ \midrule
        \multirow{4}{*}{Kodak24} & CBM3D & 35.33 & 34.28 &30.21 &32.32 &31.68 &29.68 &30.53&29.90&28.52& 28.81&28.46&28.12& 27.16& 26.82&26.66\\
        & CDnCNN & 35.23& 34.48&29.21&32.28 & 32.03 &28.93 & 30.57 &30.46 &28.26 &28.98&28.85&27.76 &25.00&25.04&24.69\\
        & FFDNet & 35.10 & 34.63 & 30.47 &32.26 & 32.13 &29.81 &30.59&30.57&29.15 & 28.98 &28.98& 28.22&27.26 &27.27 &26.92 \\
         & \cellcolor{Gray}FINO & \cellcolor{Gray}\textbf{35.41} & \cellcolor{Gray}\textbf{34.67} & \cellcolor{Gray}\textbf{30.68}&\cellcolor{Gray}\textbf{32.57} &\cellcolor{Gray}\textbf{32.31} &\cellcolor{Gray}\textbf{30.10} &\cellcolor{Gray}\textbf{29.21} &\cellcolor{Gray}\textbf{30.62} &\cellcolor{Gray}\textbf{29.31} &\cellcolor{Gray}\textbf{29.35} &\cellcolor{Gray}\textbf{29.14} &\cellcolor{Gray}\textbf{28.45} & \cellcolor{Gray}\textbf{27.65} & \cellcolor{Gray}\textbf{27.54}& \cellcolor{Gray}\textbf{27.33} \\
\bottomrule
    \end{tabular}
}
\label{tab:gau_res}
\end{table*}

\vspace{1mm}
\noindent\textbf{Noise correlation matrix.}
We assume that the additive noise $\textbf{n}$ is spatially uniform and uncorrelated (\eg \textit{i.i.d.} Gausian noise).
Let $V:\mathbf{n}\mapsto V\mathbf{n}\in \mathbb{R}^{m\times M}$ be an overlapping patch extractor, where $m$ denotes the number of pixels within one patch and $M$ denotes the number of patches.
We obtain the noise patch matrix $\hat{\mathbf{N}} = V\hat{\mathbf{n}} = [\hat{\mathbf{N}}_1, \hat{\mathbf{N}}_2, \ldots, \hat{\mathbf{N}}_M]$ from the reconstructed $\hat{\mathbf{n}}$.
The patch-wise noise correlation matrix is defined as
\begin{equation}\begin{aligned}
\mathbf{\Sigma} = \frac{1}{M} \sum^M_{j=1}\hat{\mathbf{N}}_j\hat{\mathbf{N}}_j^T\;.
\end{aligned}\end{equation}

Based on the uncorrelated noise assumption, all of the non-diagonal elements of $\mathbf{\Sigma}$ should be as close to zero as possible. Denote the standard deviation of $\textbf{n}$ to be $\sigma$, the diagonal elements of $\mathbf{\Sigma}$ should all be $\sigma^2$. Therefore, we set the following \textit{noise correlation loss} as
\begin{align}
\mathcal{L}_{noise} = \| \mathbf{\Sigma} - \sigma^2\mathbf{I}\|^2_F\;.
\end{align}




\vspace{1mm}
\noindent\textbf{Full objective function.}
By combining the above losses, the hybrid objective function $\mathcal{L}$ used to train our model is
\begin{align}
\mathcal{L} = \underbrace{\mathcal{L}_{reg}}_{\text{image}} + \underbrace{\alpha\mathcal{L}_{rec} + \beta\mathcal{L}_{cnt}}_ {\text{disentangling}}
+ \underbrace{\gamma\mathcal{L}_{noise}}_{\text{noise}} \;,
\end{align}
where $\alpha$, $\beta$, and $\gamma$ are the weighting coefficients to balance the influence of each term.

\section{Experiments}
\begin{figure}[!t]
    \centering
    \begin{minipage}{1.\linewidth}
        \centering
        \begin{subfigure}[b]{0.49\textwidth}
            \centering
            \includegraphics[width=\textwidth]{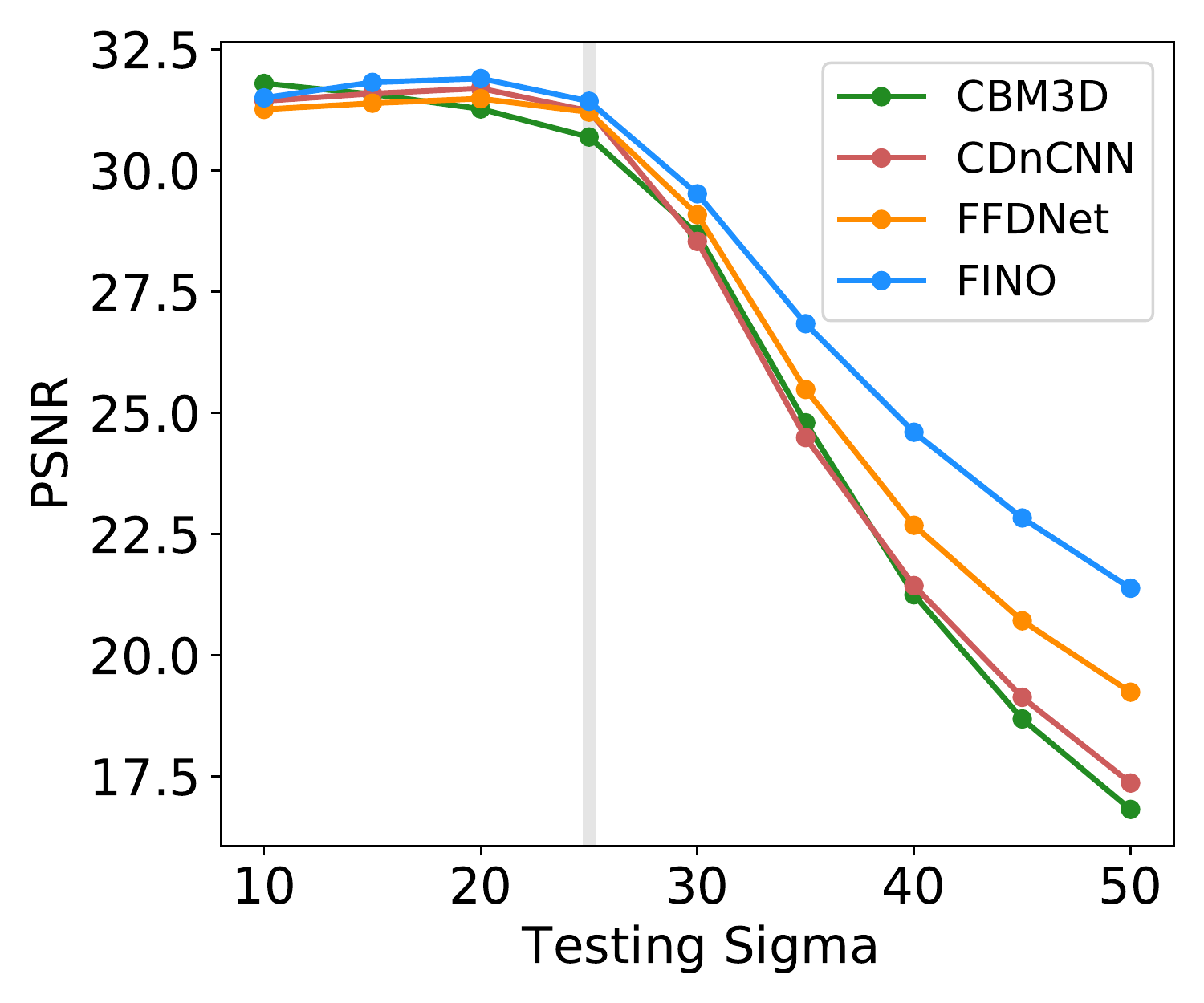}
            \label{fig:observation1}
        \end{subfigure}
        \begin{subfigure}[b]{0.49\textwidth}
            \centering
            \includegraphics[width=\textwidth]{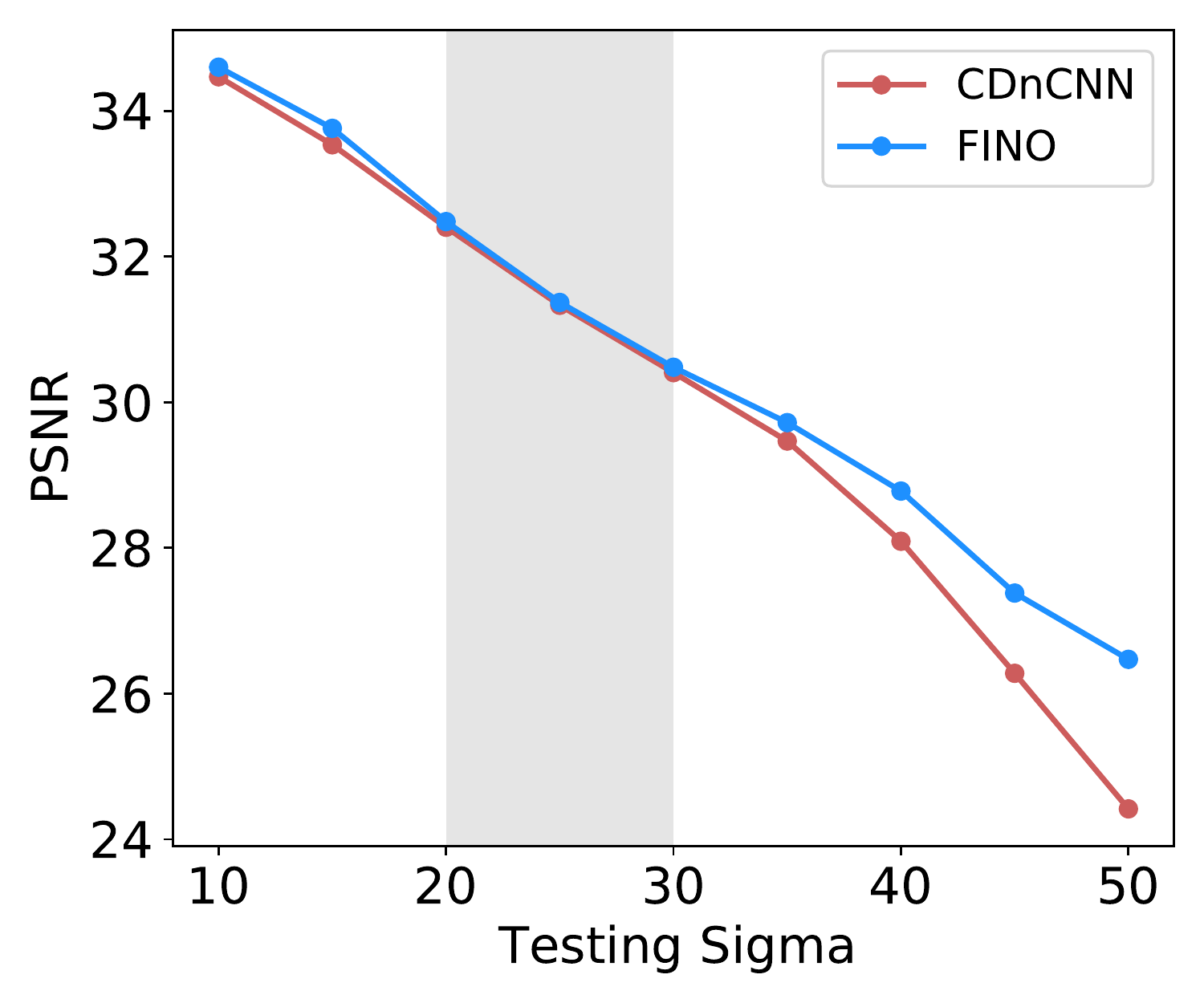}
            \label{fig:observation2}
        \end{subfigure}
        \vspace{-0.5cm}
          \caption{The generalization capability of the image denoising methods. All the methods are trained over a fixed ranges of noise levels which are indicated by the gray intervals.
        }
        \label{fig:observation}
    \end{minipage}
\end{figure}

\begin{figure*}[t]
\centering
\includegraphics[width=1.\linewidth]{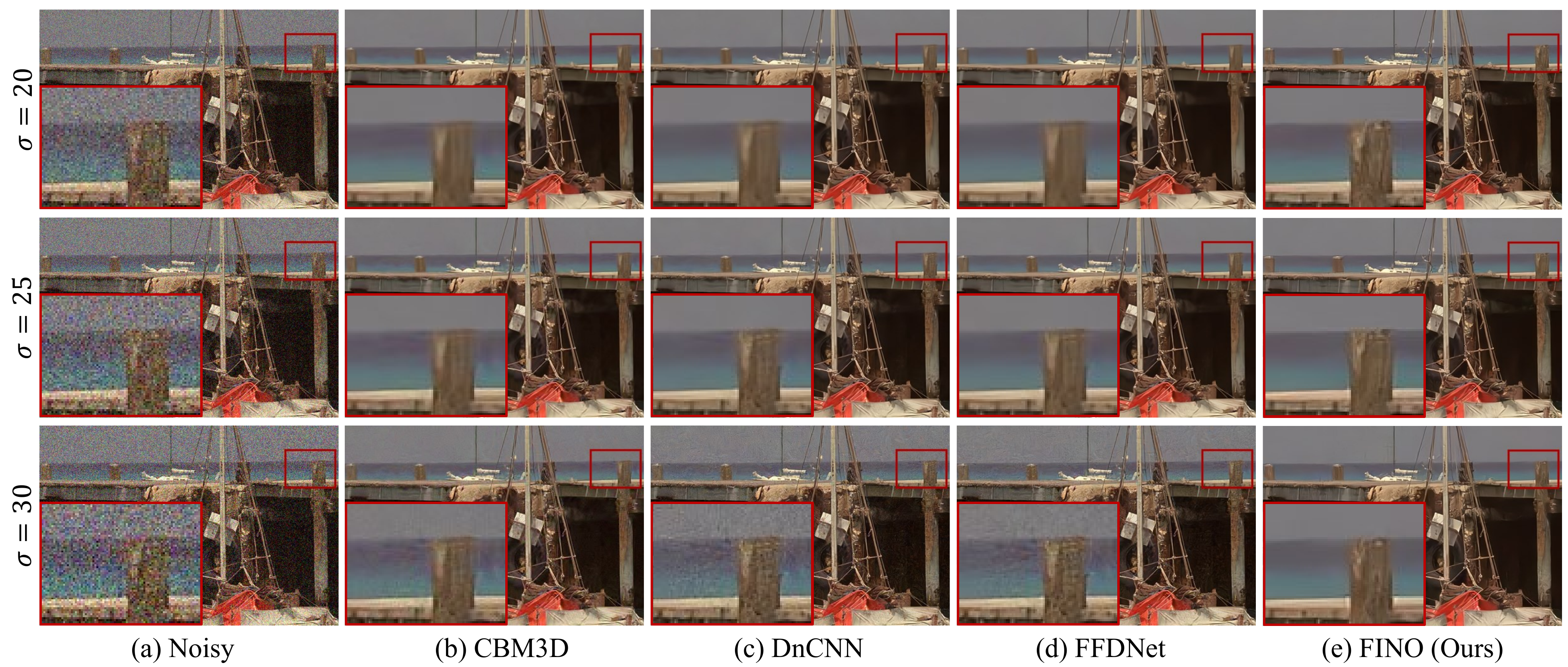}
\vspace{-3mm}
\caption{
Examples of the denoised results on Kodak24~\cite{franzen1999kodak} dataset. From left to right, the input noisy image, the estimated results of CBM3D~\protect\cite{dabov2007color}, CDnCNN~\protect\cite{zhang2017beyond}, FFDNet ~\protect\cite{zhang2018ffdnet}, and our method.
All the methods are set/trained with the estimated noise level $\sigma=25$. From top to down, the results for noisy images with $\sigma=15$, $\sigma=25$, and $\sigma=35$.
}

\label{fig:gau_vis}
\end{figure*}

We evaluate the denoising performance of the proposed FINO on both synthetic and real noise with extensive experiments. FINO is implemented using PyTorch, which is tested on a GTX 2080Ti GPU.
We adopt the ADAM optimizer with an initial learning rate of $4\times 10^{-4}$. 
FINO employs two Flow Blocks and twelve coupling layers in each block.
The ratio of clean image and noise variables is fixed as $3:1$, since content contains richer texture and edge information.
We set the hyper-parameters $\alpha=1$, $\beta = 1$, and $\gamma=0.1$.
The network parameters are initialized randomly. During training, we randomly crop patches of resolution $144\times 144$ from input images.
We employ Peak Signal-to-Noise (PSNR) for the quantitative evaluation of denoised results.
Since the real noisy degradation is relatively mild, causing small PSNR differences in some cases, we also report the Structural SIMilarity (SSIM) for real noise removal experiments.

\subsection{Synthetic Noise Removal}
We simulate spatially invariant additive white gaussian noise (AWGN) with different $\sigma$ to evaluate the synthetic noise removal performance. We also evaluate if the denoising methods can be generalized to $\sigma$ that is different from the training corpus.
Besides the uniform noise, we further simulate spatially variant AWGN to evaluate the method's robustness.

\begin{figure}[t]
	\begin{center}
			\includegraphics[width=1.\linewidth]{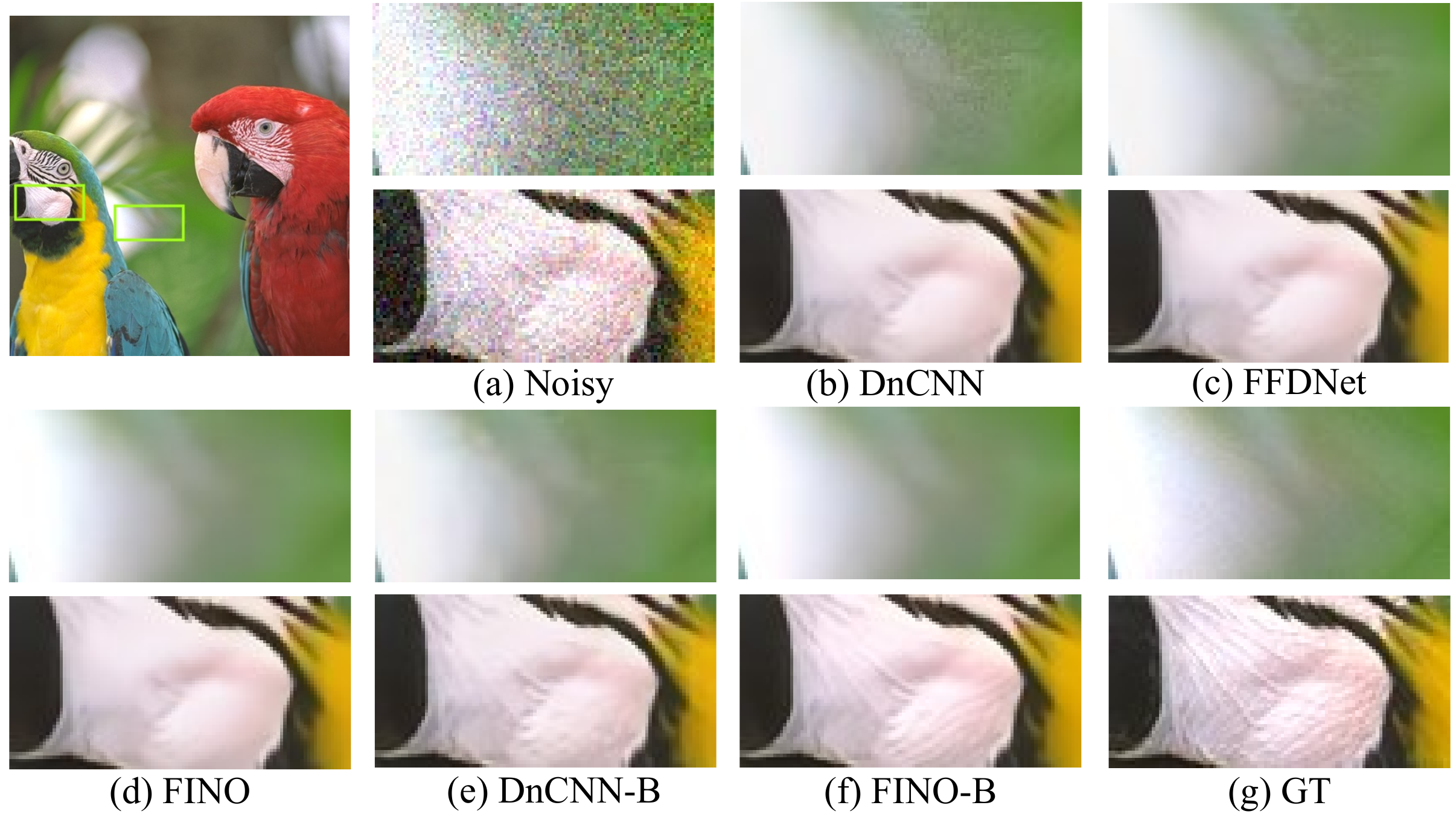}
	\end{center}
	\vspace{-0.3cm}
	\caption{Generalization capacity of different methods on spatially variant noise removal ($\sigma \in (15,35)$). All the methods are set/trained with the estimated noise level $\sigma=25$, except to the blind denoising model, \ie CDnCNN-B and FINO-B. }
	\label{fig:nonuniform}
\end{figure}

\vspace{1mm}
\noindent\textbf{Spatially invariant AWGN.}
We evaluate the proposed method in AWGN removal on two widely-used image denoising datasets: CBSD68~\cite{roth2005fields} and Kodak24~\cite{franzen1999kodak}.
The CBSD68 dataset consists of 68 images from the separate testing set of the BSD300 dataset~\cite{roth2005fields}. 
The Kodak24 dataset consists of 24 center-cropped images of size $500\times 500$ from the original Kodak dataset.
We only use 200 images selected from the training set of the BSD dataset as training data.
The noisy images are obtained by simulating AWGN of noise level $\sigma = 15,25,35,50,75$ to the clean counterpart.
We compare the proposed FINO method with several state-of-the-art denoising methods, including one widely-used classic method (\ie CBM3D~\cite{dabov2007color}), and deep learning based methods (\ie CDnCNN~\cite{zhang2017beyond} and FFDNet~\cite{zhang2018ffdnet}).
We first test FINO on noisy images corrupted by spatially invariant AWGN.
In practice, it is difficult to estimate the noise level accurately, and the estimated noise level can vary in a range. 
Most existing methods are sensitive to the estimated noise level~\cite{Mohan2020Robust}, which means the performance would severely degrade while applied to wrongly estimated noise level.
Hence, besides testing the performance when the estimated noise level is accurate, we also test the cases when the noise level is wrongly estimated.
For model trained with each estimated $\sigma$, we test their performance on $\{\sigma-5, \sigma, \sigma+5\}$ truly sigma variance. 
The competing methods also are evaluated following the same settings.
From the quantitative results shown in Table~\ref{tab:gau_res}, our method outperforms all competing methods, especially for the wrong $\sigma$ estimation cases.
Although some competing methods are good at removing Gaussian noise when accurate noise estimation, their performance would be significantly degraded once the noise level is wrongly estimated, even slight variation as shown in Figure~\ref{fig:observation}.
The examples in Figure~\ref{fig:gau_vis} also verify the observations in Table~\ref{tab:gau_res}.
With the merits of disentanglement and noise model, FINO has strong generalization capability comparing to other methods.

\begin{table}[!t]
\centering
\footnotesize
\setlength{\tabcolsep}{0.3em}
\caption{Quantitative results on the spatially variant noisy images on Kodak24~\cite{franzen1999kodak} dataset. All the methods are set/trained with the estimated uniform noise level $\sigma=25$, while applying to spatially variant noisy images in the testing stage. Except to the blind denoising methods CDnCNN-B and FINO-B, since they need not estimated noise level.}
\adjustbox{width=0.9\linewidth}{
    \begin{tabular}{l|cc}
        \toprule
         Method & $\sigma \in (0, 50)$ & $\sigma \in (15, 35)$ \\\midrule
         Noisy & 20.66 & 20.58 \\
        CBM3D~\cite{dabov2007color} & 29.66 & 31.16 \\
         CDnCNN~\cite{zhang2017beyond} &  29.52  & 31.41  \\ 
                  FFDNet~\cite{zhang2017beyond} & 30.00  & 31.42  \\ 
          \rowcolor{Gray}FINO (Ours) & \textbf{31.15} & \textbf{32.01}  \\
          \midrule
        CDnCNN-B~\cite{zhang2017beyond} &  31.88 & 31.75  \\ 
                    \rowcolor{Gray}FINO-B (Ours) & \textbf{32.43} & \textbf{32.36}  \\
        \bottomrule
    \end{tabular}
}
\label{tab:nonuniform}
\end{table}

\vspace{1mm}
\noindent\textbf{Spatially variant AWGN.}
We further evaluate the generalization capability of the proposed FINO to deal with spatially variant AWGN.
We follow the spatially variant AWGN synthetic approach in~\cite{zhang2018ffdnet}, which generates a noise level map and applies it to images using element-wise multiplication.
We select classic methods (CBM3D~\cite{dabov2007color}) and deep learning based methods (CDnCNN~\cite{zhang2017beyond}, FFDNet~\cite{zhang2018ffdnet}, and CDnCNN-B (blind version of CDnCNN) as the competing methods.
In this experiment, we evaluate two versions of FINO: (1) FINO is trained on noisy images with specific $\sigma=25$; (2) FINO-B is trained on noisy images with a $\sigma$ range of $(0,55]$.
In the denoising stage, the ground truth noise level map is unavailable. All the methods are set/trained with the estimated noise level $\sigma=25$, except to the CDnCNN-B, which is trained over a range of noise level $(0,55]$ and does not need an estimated noise level.
The quantitative results are shown in Table~\ref{tab:nonuniform}. Our method outperforms all competing methods by large margins, and our blind denoising version also achieves better performances compared with CDnCNN-B.
The examples shown in Figure~\ref{fig:nonuniform} demonstrates our method can coarsely estimate the spatial noise level and effectively reconstruct the clean components, while the other competing methods deliver more over-smoothness or noisy residuum.

\begin{table}[!t]
\centering
\footnotesize
\caption{Quantitative results on the real noisy SIDD dataset trained on SIDD medium dataset. The best result is in
\textbf{bold}.}
\adjustbox{width=0.85\linewidth}{
    \begin{tabular}{l|cc}
        \toprule
         Method & PSNR $\uparrow$  & SSIM $\uparrow$ \\\midrule
         DnCNN~\cite{zhang2017beyond} & 23.66  & 0.583 \\ 
        TNRD~\cite{chen2016trainable} & 24.73  & 0.643 \\
        CBM3D~\cite{dabov2007color} & 25.65 & 0.685 \\
        CBDNet\cite{guo2019toward} &33.28  & 0.868  \\
        GradNet~\cite{liu2020gradnet} & 38.34 & 0.953 \\
        AINDNet~\cite{kim2020transfer} & 39.08  & 0.955   \\
        VDN~\cite{yue2019variational} &39.26 & 0.955  \\
          InvDN~\cite{liu2021invertible} & 39.28 &  0.955  \\
          \rowcolor{Gray}FINO (Ours) &  \textbf{39.40} &\textbf{ 0.957 } \\
        \bottomrule
    \end{tabular}
}
\label{tab:real}
\end{table}

\subsection{Real RGB Noise Removal}
Finally, we evaluate the performance of different methods on a real-world dataset, which follows a more complicated noise distribution.
In this experiment, we drop the noise modeling loss $\mathcal{L}_{noise}$, since the real noise in SIDD dataset seems against the uncorrelated noise assumption.
Real noise stems from multiple sources, \eg short noise, thermal noise, and dark current noise, and is further affected by the in-camera processing (ISP) pipeline, which can be much more different from uncorrelated noise.
We conduct real noise removal on the SIDD dataset~\cite{SIDD_2018_CVPR}, which is taken by five different types of smartphones. We utilize the medium SIDD dataset as the training set, which contains 320 clean and noise image pairs.
We select classic denoising algorithm, CBM3D~\cite{dabov2007color}, one blind Gaussian denoising method CDnCNN-B~\cite{zhang2017beyond}, and real denoising methods, \ie CBDNet~\cite{guo2019toward}, GradNet~\cite{liu2020gradnet}, VDN~\cite{yue2019variational}, InvDN~\cite{liu2021invertible}, as well as synthetic denoiser transferring to real noise, \ie AINDNet~\cite{kim2020transfer}.
The performance comparison on the test set of the SIDD dataset is listed in Table~\ref{tab:real}.
The proposed FINO outperforms all competing methods. 
With the merits of the great generalization capability, the FINO can perform blind real image denoising without an external noise estimation module.
Besides, the number of parameters of the FINO is (3.96M), which is much smaller than some competing methods, such as AINDNet (13.76M) and VDN (7.81M).

\begin{figure}[t]
	\begin{center}
			\includegraphics[width=1.\linewidth]{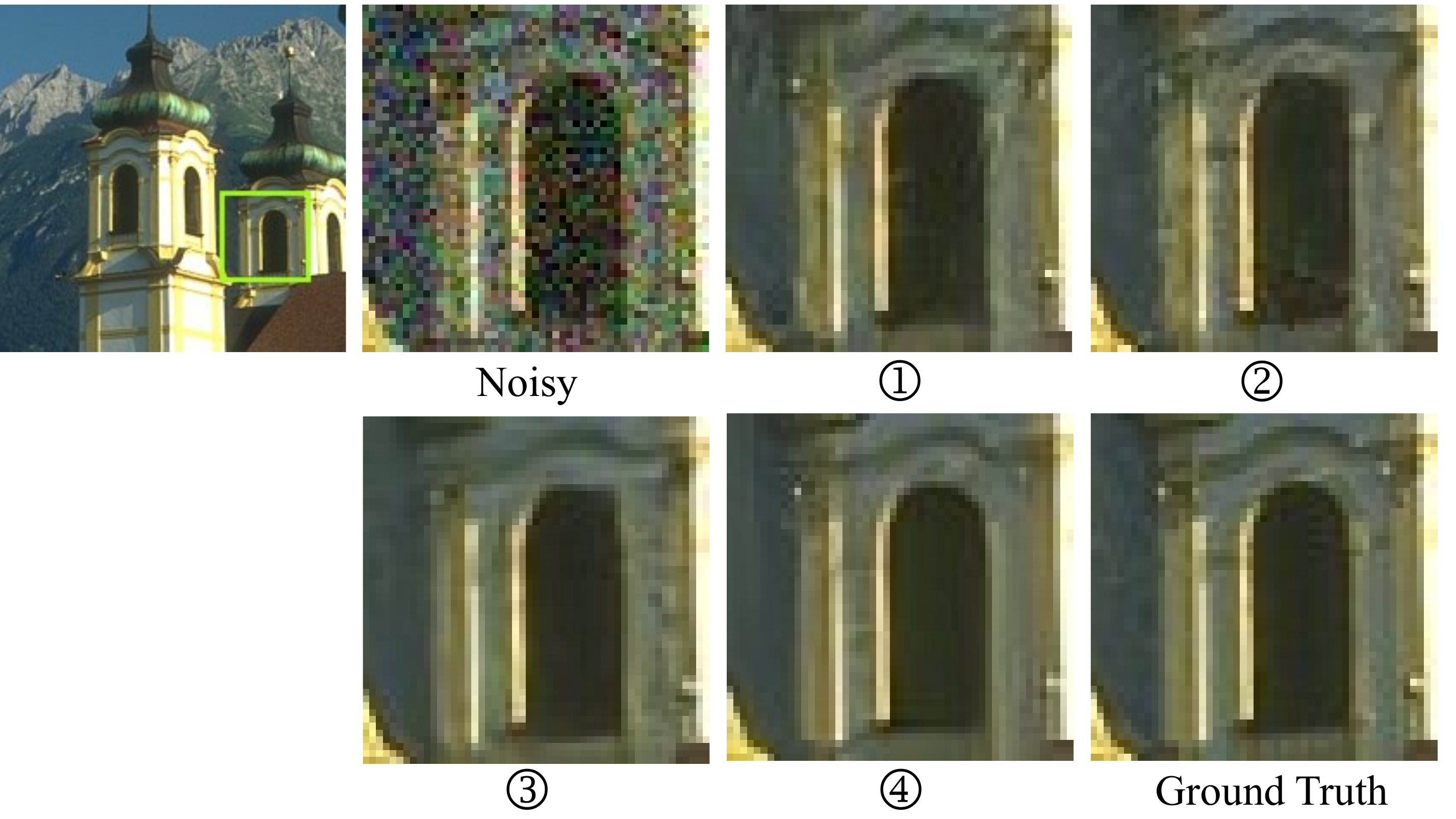}
	\end{center}
	\vspace{-2mm}
	\caption{Visual examples of ablation study, including noisy image, results of four ablation experiments corresponding to the No. in Table~\ref{tab:ablation}, and ground truth.}
	\label{fig:ablation}
\end{figure}

\begin{table}[!t]
\centering
\footnotesize
\caption{Quantitative results of ablation study. Evaluated model is trained on $\sigma=25$ and applied to $\sigma=20,25,30$ for generalization capacity measurements. Note that all variant models include the regression loss $\mathcal{L}_{reg}$.}
\adjustbox{width=0.8\linewidth}{
    \begin{tabular}{c|ccc|ccc}
        \toprule
      \multirow{2}{*}{} & \multirow{2}{*}{$\mathcal{L}_{cnt}$} & \multirow{2}{*}{$\mathcal{L}_{rec}$} & \multirow{2}{*}{$\mathcal{L}_{noise}$} &\multicolumn{3}{c}{$\sigma=25$} \\
        & & & & 20 & 25 & 30 \\ \midrule
    \ding{172} & & \Checkmark& &31.67 &31.20 & 29.17 \\
       \ding{173} &\Checkmark & 
       & & 31.66 & 31.24 & 29.34  \\
        \ding{174} &\Checkmark & \Checkmark & & 31.80 & 31.32 & 29.47  \\
      \rowcolor{Gray} \ding{175} & \Checkmark & \Checkmark & \Checkmark & \textbf{31.82} &\textbf{31.43} &\textbf{29.52}\\

        \bottomrule
    \end{tabular}
}
\label{tab:ablation}
\end{table}

\begin{figure}[t]
	\begin{center}
			\includegraphics[width=1.\linewidth]{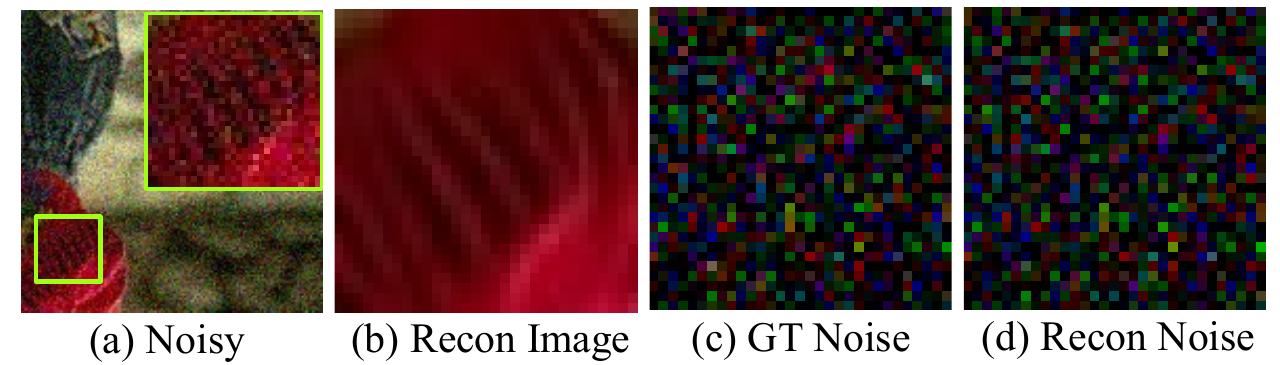}
	\end{center}
	\vspace{-0.2cm}
	\caption{Visualization of reconstructed image and noise.
	We amplified the real noise (c) and reconstructed noise (d) with $2\times$ for better visibility.
	}
	\label{fig:visulization}
\end{figure}

\subsection{Ablation Study}
Furthermore, we thoroughly investigate the impact of each loss function applied in the training stage.
Table~\ref{tab:ablation} shows the evaluation results and Figure~\ref{fig:ablation} demonstrates the visual examples on different combination of loss functions.
Visual quality of results by FINO without content alignment loss and reconstruction loss drop significantly as shown in the \ding{172} and \ding{173} in Figure~\ref{fig:ablation}.
Without the content alignment loss, the generalization capability decreases largely, especially for the robustness for the higher noise as shown in the \ding{172} in Table~\ref{tab:ablation}.
Without the reconstruction loss, the generalization capability would decrease since lacking a strong disentangling constraint.
The visual example in Figure~\ref{fig:ablation} also verifies the result with noise loss preserves more structural details.

\subsection{Visualization of Noise Disentanglement}
To better understand the noise model in FINO, we visualize the reconstructed noise and image outputs in Figure~\ref{fig:visulization}(b) and (c), respectively. We observe that the reconstructed noise is highly consistent with the ground truth one, demonstrating the effectiveness of FINO's noise modeling.

\section{Conclusion}
In this work, we propose a Flow-based joint Image and NOise model (FINO) to tackle image denoising problems, which aim to estimate the underlying clean image from its noisy measurements. FINO distinctly decouples the image and noise components in the latent space and re-couples them via invertible transformations. Based on that, we employ joint image and noise modeling, \ie image priors can be learned from the noise-free training corpus, and the noise components are modeled based on the uncorrelated noise assumption.
Our experimental results show promising results on both synthetic noise and real noise removal.
Furthermore, we demonstrate that our method has superior generalization capability to the removal of non-uniform noise and noise with inaccurate estimation.


\bibliography{aaai22}

\end{document}